\def\year{2021}\relax
\documentclass[letterpaper]{article} 
\usepackage{multirow} 
\usepackage{amsfonts} 
\usepackage[switch]{lineno}  
\usepackage{aaai21}  
\usepackage{times}  
\usepackage{helvet} 
\usepackage{courier}  
\usepackage[hyphens]{url}  
\usepackage{graphicx} 
\usepackage{natbib}  
\usepackage{caption} 
\title{Full Matching on Low Resolution for Disparity Estimation}
\author {
        Hong Zhang\textsuperscript{\rm 1}
        Shenglun Chen \textsuperscript{\rm 1}
        Zhihui Wang \textsuperscript{\rm 1} 
       Haojie Li \textsuperscript{\rm 1} 
        Wanli Ouyang \textsuperscript{\rm 2} \\
}
\affiliations {
    \textsuperscript{\rm 1} Dalian University of Technology \\
    \textsuperscript{\rm 2} The University of Sydney \\
   \{ jingshui, 1936902534\}@mail.dlut.edu.com, \{zhwang, hjli\}@dlut.edu.cn, wanli.ouyang@sydney.edu.au
}
\begin{document}

\maketitle

\begin{abstract}
A Multistage Full Matching disparity estimation scheme (MFM) is proposed in this work. We demonstrate that  decouple all similarity scores directly from the low-resolution 4D volume step by step instead of estimating low-resolution 3D cost volume through focusing on optimizing the low-resolution 4D volume iteratively leads to more accurate disparity. To this end, we first propose to decompose the  full matching task into multiple stages of the cost aggregation module. Specifically, we decompose the high-resolution predicted results into multiple groups, and every stage of the newly designed cost aggregation module learns only to estimate the results for a group of points. This alleviates the problem of feature internal competitive when learning similarity scores of all candidates from one low-resolution 4D volume output from one stage. Then, we propose the strategy of \emph{Stages Mutual Aid}, which takes advantage of the relationship of multiple stages to boost similarity scores estimation of each stage, to solve the unbalanced prediction of multiple stages caused by serial multistage framework. Experiment results demonstrate that the proposed method achieves more accurate disparity estimation results and outperforms state-of-the-art methods on Scene Flow, KITTI 2012 and KITTI 2015 datasets.
\end{abstract}

\section{Introduction}
Stereo matching\cite{DBLP:journals/pami/SunZS03,Ke2009Cross} is a core technique in many 3D computer vision applications, such as autonomous driving, robot navigation, object detection and recognition\cite{DBLP:conf/nips/ChenKZBMFU15,DBLP:conf/iccv/ZhangLCCCR15}. It aims to get the depth information for the reference image by calculating the disparity map of the two images (\emph{i.e.} left image and right image, sized $H\times W$, respectively) captured by the stereo camera. The reference image can be either left image or right image. In the rest of this manuscript, we assume the left image to be the reference image, and the right image is regarded as the source image accordingly. 

The disparity of a target point in the reference image is the horizontal displacement between the target point and its most similar point in the source image\cite{DBLP:journals/ijcv/ScharsteinS02,DBLP:conf/cvpr/LuoSU16,DBLP:conf/cvpr/SekiP17} . In order to find the most similar point for a target point, the similarity scores between this point and the candidate points in the source image are calculated(\emph{i.e.} similarity distribution)\cite{DBLP:conf/cvpr/Hirschmuller05,DBLP:journals/pami/BrownBH03}. When there are $D_{max}$ candidate points(\emph{i.e.} matching range), a 3D cost volume of size $D_{max}\times H \times W$ containing all similarity scores for $H \times W$ target points is calculated. 

To obtain such 3D cost volume, recent cost aggregation network based learning methods\cite{DBLP:conf/iccv/KendallMDH17,DBLP:conf/cvpr/ChangC18,DBLP:conf/cvpr/GuoYYWL19,DBLP:conf/cvpr/ZhangPYT19} first form a 4D volume of size $F\times \frac{1}{n}D_{max}\times \frac{1}{n}H \times \frac{1}{n}W$(where $F$ is the dimension of the correlation feature and $n$ is the ratio of downsampling) by associating each unary feature with their corresponding unary from the opposite source image across $ \frac{1}{n}D_{max}$ disparity levels. They obtain a high quality low-resolution 3D cost volume through focusing on optimizing the low-resolution 4D volume by cost aggregation network, and then get the high precision performance on final high-resolution disparity map ($H\times W$). The process of most cost aggregation networks contains multiple identical 4D volume aggregation stages to refine the correlation features multiple times. These methods only output low-resolution 3D cost volume containing similarity scores of partial candidates. To obtain high-resolution disparity,  the widely accepted method is to use the linear interpolation to get the  complete 3D cost volume firstly.  However, the similarity score is not a linear function of the disparity, which causes inaccurate estimation of the final disparity. Although some methods add additional refinement modules to refine the disparity, they still cannot get satisfactory results due to the lack of correlation features between matching pairs.
 
Actually, due to the nature of CNN, each point in the low resolution features contains the information of all pixels in the patch of the original resolution images where its located.  Therefore, all $D_{max}$ correlation features between target points and candidates are included in the low-resolution 4D volume. Leveraging convolutional layers for decoupling all $D_{max}$ similarity scores from the 4D volume is naturally a better solution for complete 3D cost volume. Early methods, such as GC-Net\cite{DBLP:conf/iccv/KendallMDH17}, decouple all $D_{max}$ similarity scores by \emph{Transposed convolution}. However, the implementation of \emph{Transposed convolution} introduces additional computation. More notably, as the network deepens, details in 4D Volume will be lost. In addition, learning $D_{max}$ similarity scores from the optimized low scale 4D volume, containing $\frac{1}{n}D_{max}$ correlation features for each target point, means that one correlation feature outputs $n$ similarity scores.  This is an internal competitive task, because each feature essentially represents the degree of the correlation between the target point and $n$ different candidates. It is too difficult for the network to compute a universal correlation features to predict $n$ optimal similarity scores of $n$ different candidates simultaneously.   

Based on the above analysis, we design a new Multistage Full Matching scheme (MFM) in this work through simply decomposing the full matching task into multiple stages. Each stage estimate a different $\frac{1}{n}D_{max}$ similarity scores. In this decomposing way, we can not only learning all similarity scores directly from the low-resolution 4D volume, but also keep one similarity score learning from one correlation feature. 

While it is noteworthy that we share the similar insight with the existing multistage matching methods\cite{DBLP:conf/cvpr/TonioniTPMS19,DBLP:conf/iccv/DuggalWMHU19,DBLP:conf/cvpr/YangMAL20,DBLP:conf/cvpr/YinDY19}, as decomposing the matching task into multiple stages. Such methods first obtain a coarse disparity and then perform residual disparity search from the neighbor of the current disparity by constructing a partial cost volume. The later stage strongly depend on  the previous stage. In contrast, the previous stage only provide a reference for the later stage in MFM. 

Multiple tasks in the proposed MFM are equally important. However, serial multistage framework, which is designed for more sufficient cost aggregation, results in unbalanced prediction of multiple stages. Aiming at this problem, we propose the strategy of \emph{Stages Mutual Aid}. Specifically, we take advantage of the close distribution of the similarities predicted at each stage, and merge the output of other $(n-1)$ stages to obtain a voting result of the current similarity distribution for reference in the current stage. In this way, not only can the shallower network exploit the output of the deeper network, but also the voting result provides a correction message for the current prediction. 

The contributions of our work are summarized as follows:
\begin{itemize} 
\item A Multistage Full Matching disparity estimation scheme (MFM) is proposed to decompose the full matching learning task into multiple stages and decouple all $D_{max}$ similarity scores from the low-resolution 4D volume step by step, which improves the stereo matching precision accordingly. 
\item A \emph{Stages Mutual Aid} strategy is proposed to solve the unbalance between the predictions of each stage in the serial multistage framework.
\end{itemize}
We evaluate the proposed method on three challenging datasets, \emph{i.e.} SceneFlow\cite{DBLP:conf/cvpr/MayerIHFCDB16}, KITTI 2012\cite{DBLP:conf/cvpr/GeigerLU12} and KITTI 2015 datasets\cite{DBLP:conf/cvpr/MenzeG15}. The results demonstrate that our MFM scheme achieves state-of-the-art.

\section{Related Work}
This section reviews recent end-to-end supervised deep learning stereo matching methods.
\subsection{2D CNN Regression Module Based Methods}
DispNetC\cite{DBLP:conf/cvpr/MayerIHFCDB16} is the first end-to-end trainable disparity estimation network. It forms a low-resolution 3D cost volume  by calculating  cosine distance of each unary feature with their corresponding unary from the opposite stereo image across each disparity level. Then the 3D cost volume is input to 2D CNN with left features for disparity regeression. Following DispNetC, CRL\cite{DBLP:conf/iccvw/PangSRYY17} and iRes-Net\cite{DBLP:conf/cvpr/LiangFGLCQZZ18} introduce stack refinement sub-networks to further improve the performance. SegStereo\cite{DBLP:conf/eccv/YangZSDJ18} and EdgeStereo\cite{DBLP:journals/corr/abs-1903-01700} both design multiple tasks frameworks for the disparity regression task. The former introduces semantic information in the refinement stage and the latter applies edge information in guiding disparity optimization. 

The 2D CNN regression network fails to make good use of geometric principles in stereo matching to regress accurate disparity. More recent works focus on directly optimize and compute 3D cost volume by cost aggregation networks.

\subsection{Cost Aggregation Network Based Methods}
Cost aggregation network based methods study how to optimize the low-resolution 4D volume to obtain more accurate similarity scores in the low-resolution 3D cost volume and output a better disparity map accordingly. Yu \emph{et al.}\cite{DBLP:conf/aaai/YuWWJ18} propose an explicit cost aggregation sub-network to provide better contextual information. PSMNet\cite{DBLP:conf/cvpr/ChangC18} introduces a pyramid pooling module for incorporating global context information into image features, and stacked 3D CNN hourglasses to extend the regional support of context information in cost volume. In order to make full use of the features, GwcNet\cite{DBLP:conf/cvpr/GuoYYWL19} builds the cost volume by concatenating the cost volume constructed in different ways. GANet\cite{DBLP:conf/cvpr/ZhangPYT19} proposes two new neural net layers to capture the local and the whole-image cost dependencies and to replace the 3D convolutional layer. AANet\cite{DBLP:journals/corr/abs-2004-09548} proposes a sparse points based intra-scale cost aggregation method to achieve fast inference speed while maintaining comparable accuracy. 

These methods  only output low-resolution 3D cost volume containing similarity scores of partial candidates from the iterative cost aggregation network. However, the low-resolution 3D cost volume is inadequate for calculating high-resolution disparity without correlation features.  Although the high-resolution 3D cost volume in GC-Net\cite{DBLP:conf/iccv/KendallMDH17} is obtained by \emph{Transposed convolution}, additional calculations are introduced because of the way the \emph{Transposed convolution} is implemented. In addition, it is an internal competitive task to calculate $D_{max}$ similarity features for the high-resolution 3D cost volume from only one 4D volume with $\frac{1}{n}D_{max}$ correlation features. The proposed MFM in this work decomposes the full matching task into these multiple cost aggregation stages and decouple the high-resolution 3D cost volume directly from the correlation features, which solved the aforementioned problem.

\subsection{Multistage Matching Methods}
Multistage matching methods\cite{DBLP:conf/cvpr/TonioniTPMS19,DBLP:conf/cvpr/YinDY19} first obtain a coarse disparity and then perform residual disparity search from the neighbor of the current disparity by constructing a partial cost volume. DeepPruner\cite{DBLP:conf/iccv/DuggalWMHU19}  develops a differentiable PatchMatch module that allows to discard most disparities without requiring full cost volume evaluation in the second stage. CVP-MVSNet\cite{DBLP:conf/cvpr/YangMAL20} proposes a cost volume pyramid based Multi-View Stereo Network for depth inference. 

These methods improve the stereo matching speed and avoid the process of obtaining high-resolution disparity from low-resolution 3D cost volume. However, if the cues obtained in the first stage is wrong, the subsequent fine matching will also be wrong. In contrast, the previous stage in our multistage methods has only guidance for the latter stage and does not limit the estimation range of the latter stage, which guarantees the freedom of estimation in the latter stage.

\section{Multistage Full Matching Network}

\begin{figure*}[t]
	\centering
	\includegraphics[width=0.8\textwidth]{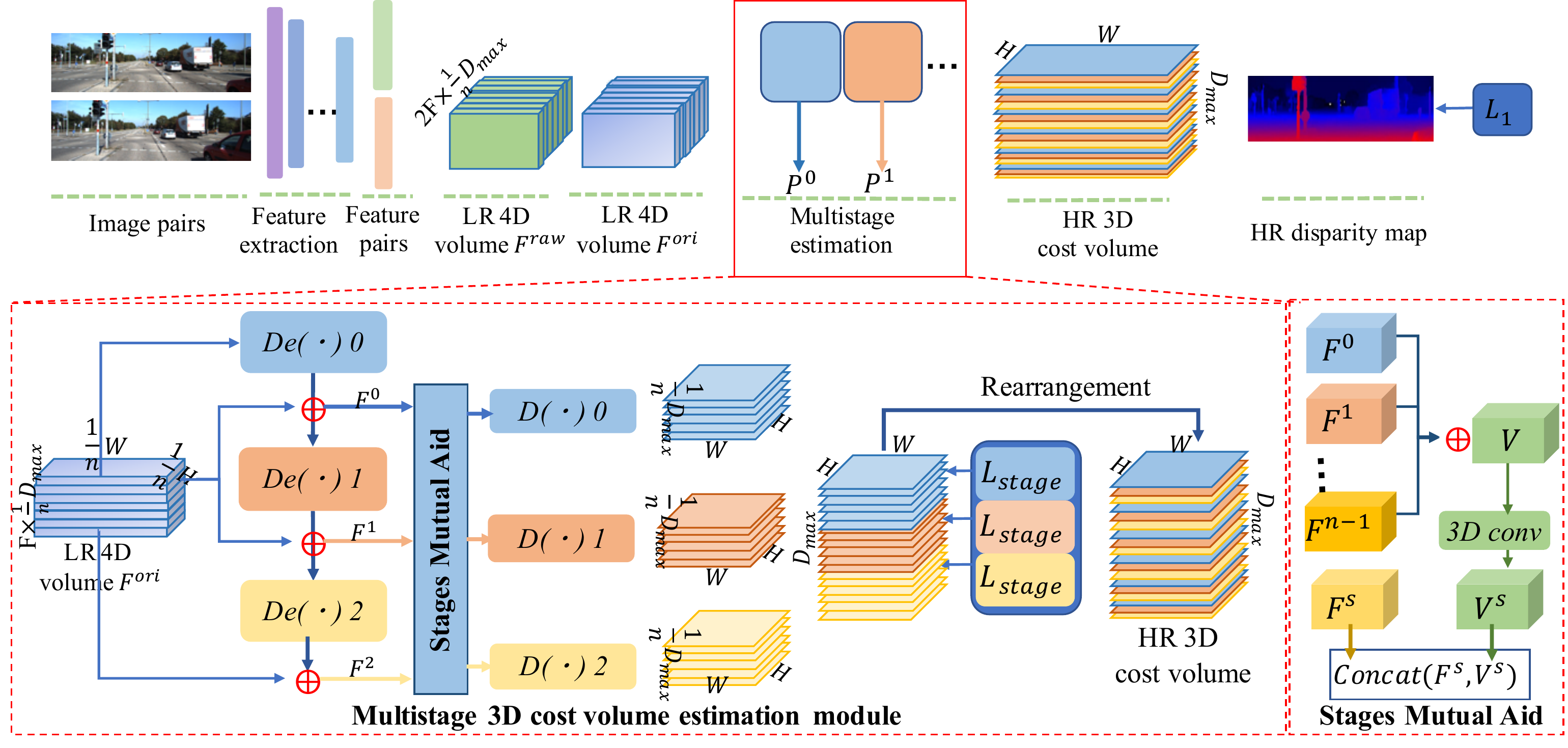}
	\caption{The architecture of the Multistage Full Matching network. LR denotes low-resolution and HR represents high-resolution. The Multistage estimation module is composed of $n$ stages. Here takes $n=3$ as an example. Each stage estimates the similarity scores of different candidate points($\frac{1}{n}D_{max}$ points) in the matching range($D_{max}$ points). The high-resolution similarity distribution($D_{max}\times H\times W$) is obtained by combining the low-resolution similarity distribution output from each stage.}
	\label{fig1}
\end{figure*}

As shown in Figure 1, the proposed framework has the following main structure: 

First, the MFM network extracts low-resolution feature maps of left and right images with shared 2D convolution. For feature extraction network, we adopt the ResNet-like network. We control the scale of the output features by controlling the stride of each convolution layer. The scale of the features is $\frac{1}{n}$ the size of the input image.  

Second, low-resolution 4D volume is calculated by \emph{Group-wise correlation}\cite{DBLP:conf/cvpr/GuoYYWL19}. The left features and the right features are divided into groups along the channel dimension, and correlation maps are computed among each group to obtain multiple cost volumes, which are then packed into one cost volume called \emph{g-cost}. Another cost volume called \emph{cat-cost} is obtained by concatenating the features of each target point and candidate points in the matching range. The final cost volume is obtained by concatenating the corresponding correlation features of \emph{g-cost} and \emph{cat-cost}.

Then, the low-resolution 4D volume from the second step is fed to the cost aggregation process for the high-resolution 3D cost volume by multistage 3D cost volume estimation module. 

At last, the final disparity is obtained by the modified parabolic fitting\cite{Raffel1998Particle,DBLP:journals/scjapan/ShimizuO02,nishiguti} sub-pixel estimation method on the high-resolution 3D cost volume.

\subsection{Multistage 3D Cost Volume Estimation Module} 
The proposed method divides the matching range into $k$ cells, and each cell contains $n$ points, where $k=D_{max}/n$. $D_{max}$ is the length of the matching range. $n$ is the same as the ratio of downsampling. In this way, the candidate set can be represented as $\{c_{0}^{0}, c_{0}^{1},c_{0}^{2}, \cdots, c_{0}^{n-1}, c_{1}^{0}, c_{1}^{1}, \cdots,  c_{m}^s, \cdots, c_{k-1}^{n-1}\}$. Each stage of the multistage full matching module learns similarity scores of specific $k$ candidates, \emph{i.e.} $\{c_{0}^{s}, c_{1}^{s}, \cdots, c_{k-1}^{s}\}$($s$ is the stage number of current stage), from $k$ correlation features of the low-resolution 4D volume. The candidates learned in the $s$-th stage is adjacent to $(s-1)$ stage and $(s+1)$ stage. $D_{max}$ high-resolution similarity scores are obtained after $n$ stages.

In order to obtain $D_{max}$ high-resolution similarity scores, we decouple $D_{max}$ similarity features from different low-resolution 4D volumes. Let $f(x)$-$C$ be the coordinate system set up with $f(x)$ as the origin in the C dimensional feature space, and $f(x)$  is the feature corresponding to the point $I_r(x)$ of reference image. The similarity features between $I_r(x)$ and its candidates, \emph{i.e.} $\{fc(x+0), fc(x+1), \cdots, fc(x+d), \cdots, fc(x+D_{max}-1)\}$, is a series points in $f(x)$-$C$. 

In camera coordinate system, the position difference between $c_m^s(x)$ can be represented as $\Delta x$. Similarly, in $f(x)$-$C$, the position difference between $fc(x+d)$ can be represented as $\Delta fc$, where $\Delta fc \in \mathbb{R}^C$. Therefore, when the position of $fc(x+d)$ is known, the position of $fc(x+d+\Delta x)$ can be obtained by:
\begin{equation}
fc(x+d+\Delta x) = fc(x+d) + \Delta fc
\end{equation}
where $\Delta x$ represents the position difference between  $c_m^s(x)$ and  $c_m^s(x+\Delta x)$.

$fc(x+d)$ in high-dimensional feature space not only contains the information of $c_m^s(x)$, but also include the features of points surround it. Consequently, $\Delta fc$ from $fc(x+d)$ to $fc(x+d+\Delta x)$ can be decoupled from $fc(x+d)$ when $\Delta x$ is small. The success of our experiment verifies the correctness of this inference.

Based on the above analysis, we design the follow framework:
\paragraph{Step 1} we first initialize a similarity feature $F^{ori}=\{fc^{ori}_m | m \in N, m < k\}$ for all stage in the space. After that, we will estimate $\Delta F^{s}=\{\Delta fc^{s}_m | m \in N, m < k\}$ from $F^{ori}$ to $F^{s}=\{fc^{s}_m | m \in N, m < k\}$ for each stage in the second step, where $s$ is the stage number.

Due to the nature of CNN, each point in the low scale features contains the information of all pixels in the patch of the original resolution images where its located.  Such features will first be associated  across $\frac{1}{n}D_{max}$ disparity levels as follows:
\begin{equation}
F_{raw}(f_{r},f_{so}^{d}) = Corr(f_{r},f_{so}^{d})
\end{equation}
where $Corr(\cdot, \cdot)$ represent \emph{Group-wise correlation}. $f_{r}$ is the feature of a target point on reference feature map, and $f_{so}^{d}$ is the feature of the $d$-th candidate point on source feature map. $F_{raw}(f_{r},f_{so}^{d})$ is the raw correlation feature of $f_{r}$ and $f_{so}^{d}$.

Therefore, all $D_{max}$  correlation features between target points and candidates are included in the raw 4D volume composed of $F_{raw}(f_{r},f_{so}^{d})$. Then, $F_{raw}(f_{r},f_{so}^{d})$ in such raw 4D volume is converted to  $F^{ori}$ by two 3D convolutions. In this way, the $F^{ori}$ can provide an excellent initial value for better $\Delta F^{s} $.

\paragraph{Step 2} The multistage similarity features estimation process is divided into $n$ stages. Except the first stage which takes $F_{ori}$ as input, each stage takes  $F^{s-1}$ as the input to get the shift  $\Delta F^{s} $ between $F^{s}$ and $F^{ori}$.
\begin{equation}
 \Delta F^{s} = De(F^{s-1} )
\end{equation}
where $De(\cdot)$ is the decouple function, which is implemented by an hourglass network composed of 3D CNN. The structure of   $De(\cdot)$ is the same as the basic block in most cost aggregation network\cite{DBLP:conf/cvpr/ChangC18,DBLP:conf/cvpr/GuoYYWL19}. Then, the simlairty features of candidates $\{c_{0}^{s}, c_{1}^{s}, \cdots, c_{k-1}^{s}\}$ in the $s$-th stage can be obtained by
\begin{equation}
F^{s} =F^{ori} + \Delta F^{s}
\end{equation}

\paragraph{Step 3} Another role of $De(\cdot)$ is to update each similarity feature by referring to the information in a larger receptive field. Therefore, a serial network composed of the $n$ $De(\cdot)$ is necessary to obtain more sufficient aggregation. However, a serial network will lead to unbalance between the predictions of each stage, for each $De(\cdot)$ is responsible for a different task. In this step,  as shown in Fig.1, a \emph{Stages Mutual Aid} operation is conducted on $F^{s}$.
 
 We design the formula(5) to construct similarity scores for supervising each stage:
\begin{equation}
S\left(m,i\right)=e^{-(m \times n+i-d_{gt})^2}
\end{equation}
where $m$ stands for the $m$-th cell, and $m \in \{0,1,2,\ldots, k-1\}$.  $n$ is the length of each cell. $i$ is the order of the candidate point in each cell, and it also represents the stage order. $d_{gt}$ is the ground-truth.  $S(m,i)$ represents the similarity score that the $i$-th point in the $m$-th cell of the $i$-th stage should have. The similarity distribution ground-truth of different stages can be obtained by changing the value of $i$. If the similarity peak falls in the $i$-th position of the $m$-th bin, the similarity peak of each stage will fall in the $m$-th bin or the ajacent bin of $m$. Therefore, the similarity distributions output of each stage is close, the similarity peak difference of which is 1 or 0. Accordingly, a voting result can be obtained from the similarity features output of the other $(n-1)$ stages for the estimation of the $s$-th stage.

First, we obtain the voting results $V$ for the $s$-th stage by 
\begin{equation}
V= \sum_{i=0, i\ne s}^{n-1}(F^i)
\end{equation}
Then, the $V$ is optimized by one 3D convolution layer, obtaining $V^s$.
Because it's hard to decide artificially the percentage of $V^s$ and $F^s$ when fusion the two features, we directly let the network learn how to merge the two features. The two features are concatenated along $d$ dimension and fed to the distance network $D(\cdot)$ to calculate the similarity scores. The $D(\cdot)$ network is consist of two 3D convolution layers.
\begin{equation}
P(I_r(x), c^s(x))) = D(Concat(V^s, F^s)), 
\end{equation}
where $P(I_r(x), c^s(x)))$ is the 3D cost volume output from the $s$-th stage, which represents the predicted similarity scores between $I_r(x)$ and $c^s(x)$ of the $s$-th stage. $Concat(\cdot)$ is the concatenation operation. 

 \paragraph{Step 4}
 Latitude $H$ and latitude $W$ of the $n$ 3D cost volumes are restored to the original scale by linear interpolation. The full resolution cost volume is obtained by combining the $n$ low-resolution 3D cost volume along the similarity dimension. Finally,  high-resolution cost volume is normalized by \emph{softmax} operation along the similarity dimension and rearranged in the similarity dimension to obtain the final high-resolution cost volume.

\subsection{Supervision Strategy and Loss Function}

We design the formula(5) to construct similarity scores for supervising each stage. By using formula(5), our supervision strategy can guarantee the relationship between the output results of multiple stages. 

The full loss of the proposed MFM network can be represented as the following:
\begin{equation}
Loss= L_{stage}+L_{1}
\end{equation}

$L_{1}$ loss is utilized to optimize the final disparity calculated through the high-resolution similarity distribution. We denote the predicted disparity as $d$ and the disparity ground-truth as $d_{gt}$, $L_{1}$ loss can be represented as the following:
\begin{equation}
L_{1} = \sum|d-d_{gt}|
\end{equation}

$L_{stage}$ is designed to guide each stage to learn specific $k$ similarity scores from $k$ correlation features of the low-resolution 4D volume. We need to align the predicted similarity distributions of $n$ stages with the aforementioned supervision similarity distributions of the $n$ stages, respectively, therefore the Cross Entropy Error Function is selected for supervision. $L_{stage}$ loss is designed as follows:
\begin{equation}
L_{stage}=\sum_{i=0}^n\sum_{m=0}^{k-1}S(m,i)\cdot logP(m,i)
\end{equation}
where $P(m,i)$ represents the similarity score between the target point and the $i$-th candidate point in the $m$-th cell output from the $i$-th stage, and $k=D_{max}/n$. 

\begin{figure*}[htb]
  \centering
  \includegraphics[scale=0.43]{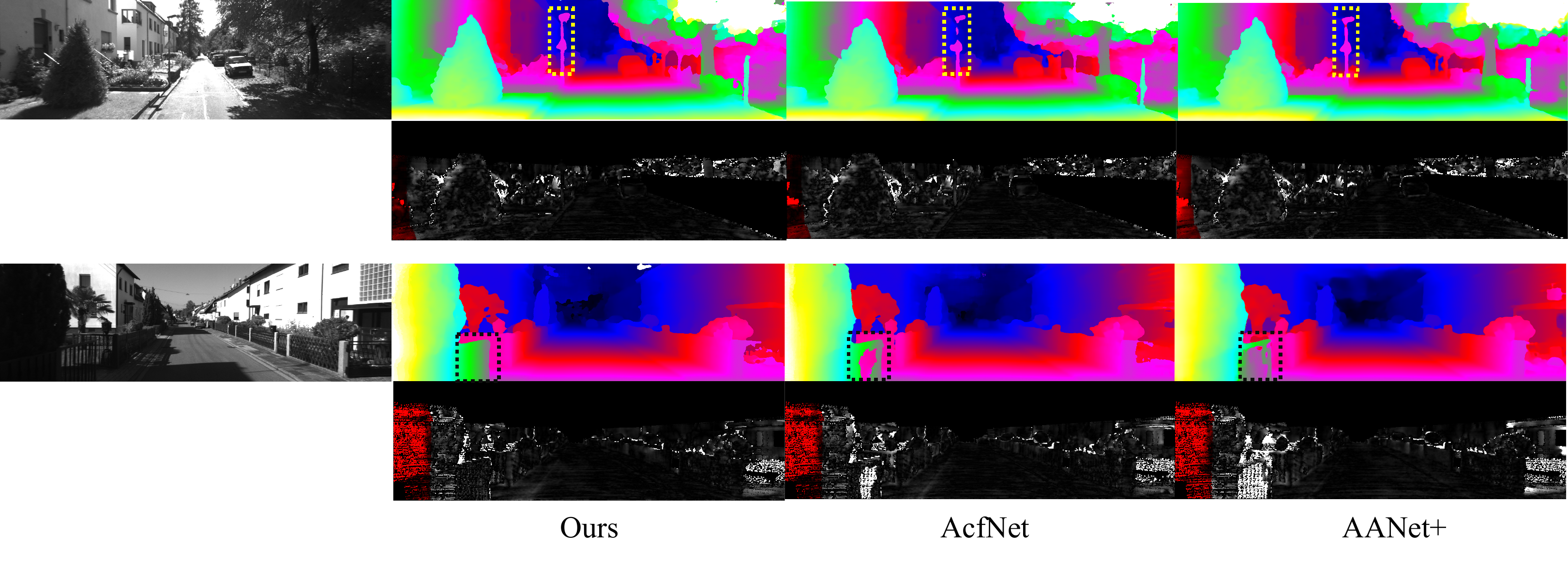}
  \caption{Error map visualization of AcfNet\cite{DBLP:conf/aaai/0005C0YYLY20}, AANet+\cite{DBLP:journals/corr/abs-2004-09548}  and our method on KITTI 2012. Darker represents lower error.}
 \end{figure*}

\begin{figure*}[htb]
  \centering
  \includegraphics[scale=0.43]{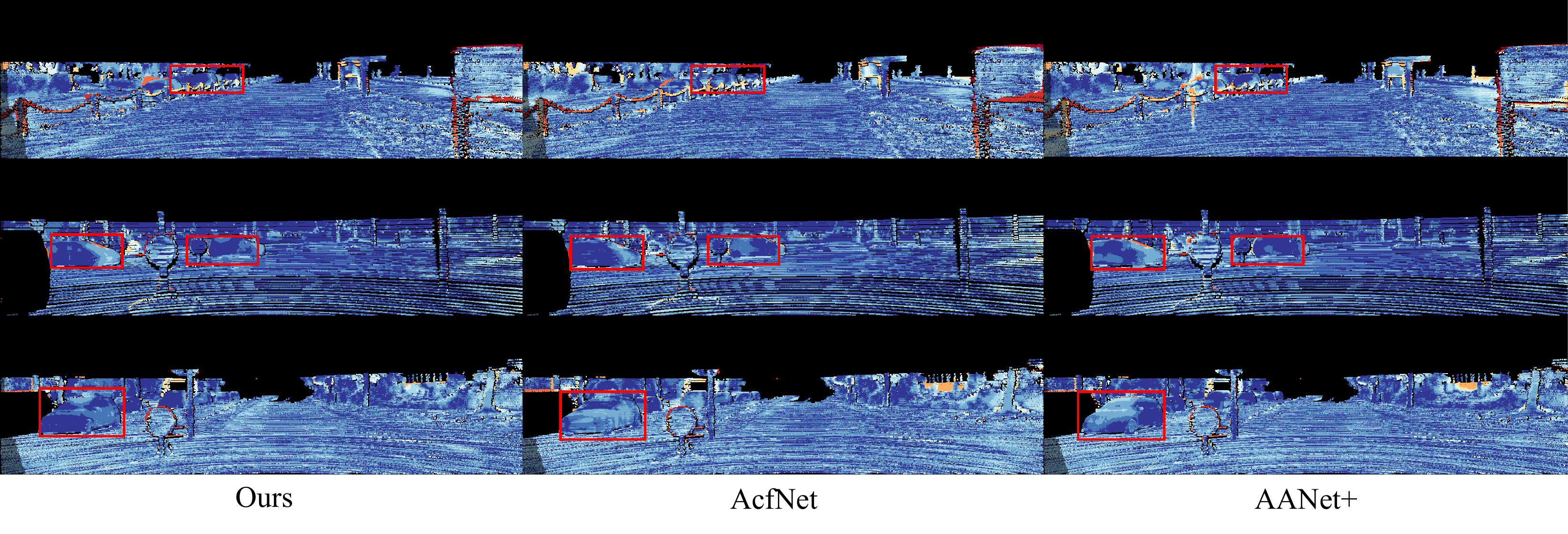}
  \caption{ Error map visualization of AcfNet\cite{DBLP:conf/aaai/0005C0YYLY20}, AANet+\cite{DBLP:journals/corr/abs-2004-09548}  and our method on KITTI 2015. Darker blue represents lower error.}
\end{figure*}
\section{Experiments}
\subsection{Implementation Details}
\paragraph{Datasets.} Scene Flow datasets\cite{DBLP:conf/cvpr/MayerIHFCDB16} provide 35,454 training and 4,370 testing images of size $960\times540$ with accurate ground-truth. We use the Finalpass of the Scene Flow datasets, since it contains more motion blur and defocus, and is more real than the Cleanpass. KITTI 2012\cite{DBLP:conf/cvpr/GeigerLU12} and KITTI 2015\cite{DBLP:conf/cvpr/MenzeG15} are driving scene datasets. KITTI 2012 contains 194 training image pairs with sparse ground truth and 195 testing image pairs with ground truth disparities held by evaluation server for submission evaluation only. KITTI 2015 contains 200 training stereo image pairs with sparse ground-truth and 200 testing image pairs with ground truth disparities held by evaluation server for submission evaluation only. 

\paragraph{Evaluation Indicators.} For Scene Flow datasets, the evaluation metrics are the end-point error (EPE), which is the mean average disparity error in pixels, and the error rates $>1px$ and $>3px$ are the percentage of pixels whose error are greater than 1 pixel and 3 pixels, respectively. For KITTI 2015, we use the percentage of disparity outliers D1 as evaluation indicators. The outliers are defined as the pixels whose disparity errors are larger than $max(3px, 0.05\cdot d_{gt})$, where $d_{gt}$ denotes the ground-truth disparity. For KITTI 2012, we use the error rates $ >2px, >3px, >4px$ and $>5px$ as evaluation indicators.

\paragraph{Training.} Our network is implemented with PyTorch. We use Adam optimizer, with $\beta_{1} = 0.9, \beta_{2} = 0.999$. The batch size is fixed to 8. For Scene Flow datasets, we train the network for 16 epochs in total. The initial learning rate is set as 0.001 and it is down-scaled by 2 after epoch 10, 12, 14. We set the $D_{max}$ as 192 and $n$ as 4, respectively. For KITTI 2015 and KITTI 2012, we fine-tune the network pre-trained on Scene Flow datasets for another 300 epochs. The learning rate is 0.001 and it is down-scaled by 10 after 210 epochs.

\subsection{Performance Comparison}

\paragraph{Quantitative Comparison.}
\begin{table}[t]
	\centering
	\caption{Performance comparison on Scene Flow datasets.}
	\begin{tabular}{lccc}
		\hline
		Model & EPE & \textgreater{}1px  & \textgreater{}3px \\ 
		\hline
		iResNet-i3\shortcite{DBLP:conf/cvpr/LiangFGLCQZZ18}&2.45&9.28\%&4.57\% \\
		CRL\shortcite{DBLP:conf/iccvw/PangSRYY17}&1.32&-&6.20\% \\
		\hline
		StereoNet\shortcite{DBLP:conf/eccv/KhamisFRKVI18}&1.10&21.33\%&8.80\%   \\
		PSMNet\shortcite{DBLP:conf/cvpr/ChangC18}&1.03& 10.32\%&4.12\%        \\
		GANet\shortcite{DBLP:conf/cvpr/ZhangPYT19}&0.81& 9.00\%&3.49\%         \\
		GwcNet\shortcite{DBLP:conf/cvpr/GuoYYWL19}&0.77&8.03\%  &3.30\%       \\
		AANet\shortcite{DBLP:journals/corr/abs-2004-09548}&0.87&9.30\%&-         \\
		AcfNet\shortcite{DBLP:conf/aaai/0005C0YYLY20}&0.87&-&4.31\%          \\   
		\hline
		DeepPruner-Best\shortcite{DBLP:conf/iccv/DuggalWMHU19}&0.86&-&- \\
		\hline
		Ours&\textbf{0.66}&\textbf{4.95\%}&\textbf{2.50\%}                    \\ 
		\hline
	\end{tabular}
    \label{table1}
\end{table}

\begin{table}[t]
	\centering
	\caption{Performance comparison on KITTI 2015 datasets.}
	\begin{tabular}{llll}
		\hline 
		Model&D1-bg&D1-fg&D1-all \\ 
		\hline
		CRL\shortcite{DBLP:conf/iccvw/PangSRYY17}&2.48\%&3.59\%&2.67\% \\
		EdgeStereo-v2\shortcite{DBLP:journals/corr/abs-1903-01700}&1.84\%&3.30\%&2.08\% \\
		SegStereo\shortcite{DBLP:conf/eccv/YangZSDJ18}&1.88\%&4.07\%&2.25\% \\
		\hline
		GCNet\shortcite{DBLP:conf/iccv/KendallMDH17}&2.21\%&6.16\%&2.87\% \\
		GwcNet-g\shortcite{DBLP:conf/cvpr/GuoYYWL19}&1.74\%&3.93\%&2.11\%   \\
		AcfNet\shortcite{DBLP:conf/aaai/0005C0YYLY20}&1.51\%&3.80\%&1.89\%   \\
		Bi3D\shortcite{DBLP:conf/cvpr/BadkiTKKSG20}&1.95\%&\textbf{3.48}\%&2.21\%         \\
		AANet+\shortcite{DBLP:journals/corr/abs-2004-09548}&1.65\%&3.96\%&2.03\% \\
		\hline
		HD\^ \ 3\shortcite{DBLP:conf/cvpr/YinDY19}&1.70\%&3.63\%&2.02\%         \\
		CSN\shortcite{DBLP:conf/cvpr/GuFZDTT20}&1.59\%&4.03\%&2.00\% \\
		\hline
		Ours&\textbf{1.51\%}&3.67\%&\textbf{1.87}\% \\
		\hline
	\end{tabular}
    \label{table2}
\end{table}
\begin{table*}[t]
	\centering
	\caption{Performance comparison on KITTI 2012 datasets.}
	\begin{tabular}{l|cc|cc|cc|cc}
		\hline
		\multirow{2}{*}{Model} & \multicolumn{2}{c|}{ \textgreater{}2px}&\multicolumn{2}{c|}{ \textgreater{}3px}&\multicolumn{2}{c|}{ \textgreater{}4px}& \multicolumn{2}{c}{ \textgreater{}5px} \\ 
		\cline{2-9} 
		&Noc&All&Noc&All&Noc&All&Noc&All \\ 
		\hline
		EdgeStereo-v2\shortcite{DBLP:journals/corr/abs-1903-01700}&2.32\%&2.88\%&1.46\%&1.83\%&1.07\%&1.34\%&0.83\%&1.04\%\\ 
		SegStereo\shortcite{DBLP:conf/eccv/YangZSDJ18}&2.66\%&3.19\%&1.68\%&2.03\%&1.25\%&1.52\%&1.00\%&1.21\%\\
		\hline
		GwcNet-gc\shortcite{DBLP:conf/cvpr/GuoYYWL19}&2.16\%&2.71\%&1.32\%&1.70\%&0.99\%&1.27\%&0.80\%&1.03\%\\ 
		GANet-deep\shortcite{DBLP:conf/cvpr/ZhangPYT19}&1.89\%&2.50\%&1.19\%&1.60\%&0.91\%&1.23\%&0.76\%&1.02\%\\ 
		AMNet\shortcite{DBLP:journals/corr/abs-1904-09099}&2.12\%&2.71\%&1.32\%&1.73\%&0.99\%&1.31\%&0.80\%&1.06\%\\ 
		AcfNet\shortcite{DBLP:conf/aaai/0005C0YYLY20}&1.83\%&2.35\%&1.17\%&1.54\%&0.92\%&1.21\%&0.77\%&1.01\%\\ 
		AANet+\shortcite{DBLP:journals/corr/abs-2004-09548}&2.30\%&2.96\%&1.55\%&2.04\%&1.20\%&1.58\%&0.98\%&1.30\%\\ 
		\hline
	        HD\^ \ 3\shortcite{DBLP:conf/cvpr/YinDY19}&2.00\%&2.56\%&1.40\%&1.80\%&1.12\%&1.43\%&0.94\%&1.19\%\\ 
		\hline
		Ours&\textbf{1.68\%}&\textbf{2.16\%}&\textbf{1.15\%}&\textbf{1.47\%}&\textbf{0.91\%}&\textbf{1.16\%}&\textbf{0.76\%}&\textbf{0.97\%}\\
		\hline       
	\end{tabular}
    \label{table3}
\end{table*}

The quantitative comparison focuses on deep learning stereo matching methods. Table 1, Table 2 and Table 3 show the performance of different methods on Scene Flow datasets,  KITTI 2012 dataset, and KITTI 2015 dataset, respectively. In each table from top to bottom, the methods are separated into four groups: (1)2D CNN regression based methods, (2)cost aggregation network based methods, (3)multistage matching methods, (4)our MFM method.

(1) methods refine the low-resolution coarse disparity with much error and noise without referring to the correlation features. Thus, such methods have lower precision. (2) methods obtain the high quality 3D cost volume iteratively from the 4D volume, which provides satisfied geometry context. Therefore, compared with (1) methods, the EPE value of (2) methods is significantly reduced to below 1.0. However, the $>1px$ error rates is still high. Because the multistage cost aggregation module only outputs a low-resolution 3D cost volume, and the high-resolution disparity is obtained from the low-resolution 3D cost volume without the geometry correlation features. (3) methods match within the narrow matching range obtained from the previous stage other than obtaining all similarity scores directly. The miscalculated narrow matching range obtained in the previous stage results in misestimating in the latter stage, which causes its low accuracy. As shown in Table1,Table2 and Table3, the proposed MFM method performs better, which demonstrates the effectiveness of the multistage 3D cost volume estimation module.
\begin{figure*}[t]
	\centering
	\includegraphics[width=0.8\textwidth]{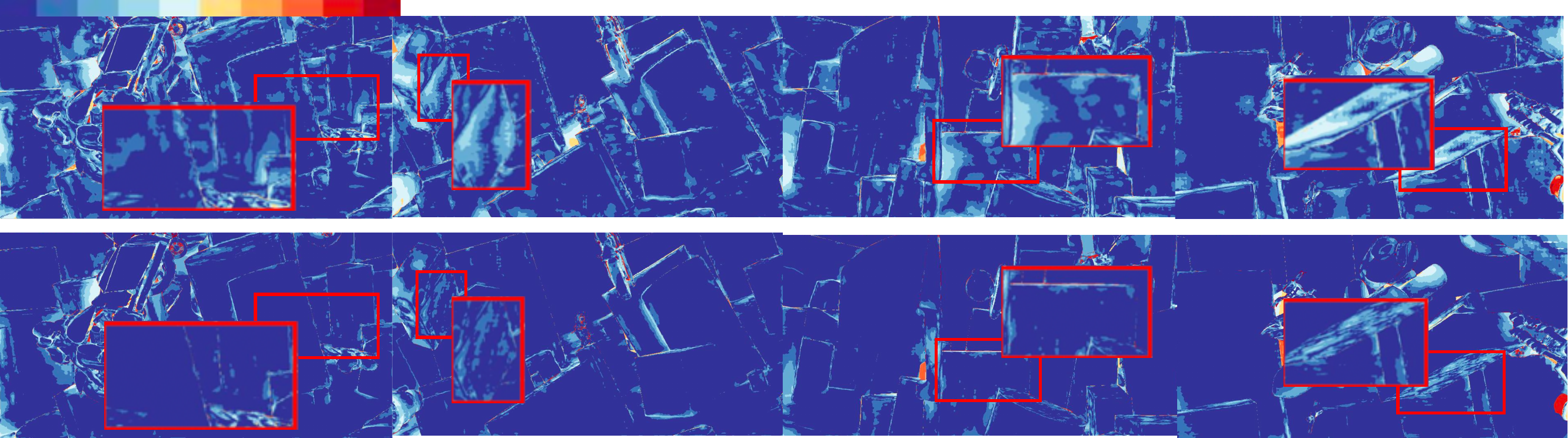}
	\caption{Error map visualization of GwcNet(top row)\cite{DBLP:conf/cvpr/GuoYYWL19} and our method(bottom row) on Scene Flow datasets. Darker blue represents lower error.}
	\label{figure2}
\end{figure*}

\paragraph{Qualitative Comparison.}
\begin{figure}[t]
	\centering
	\includegraphics[width=0.4\textwidth]{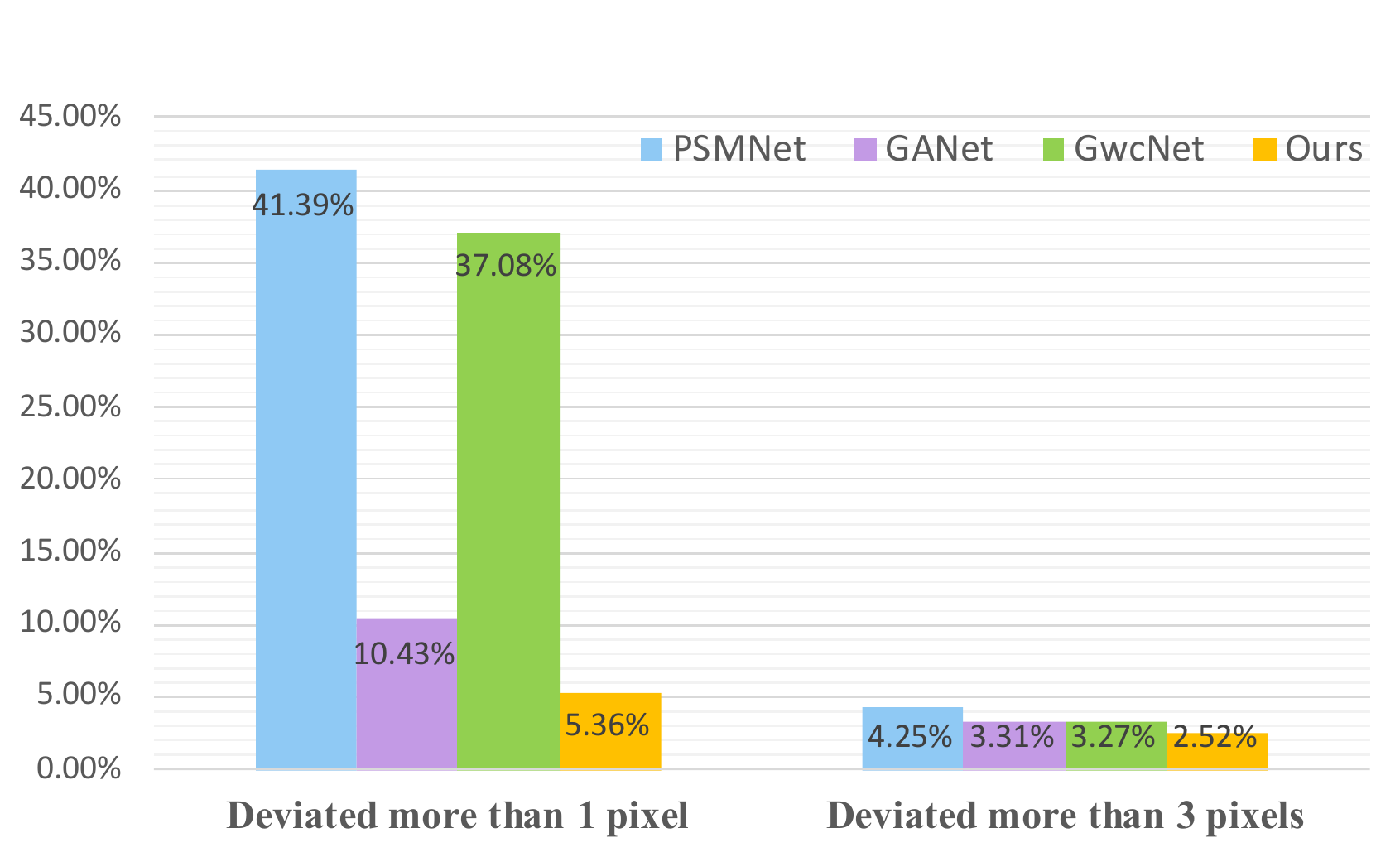}
	\caption{Peak position estimation comparison with existing classic  cost aggregation network based methods. }
	\label{figure3}
\end{figure}
Figure 4 visualizes the error maps of our method and the GwcNet\cite{DBLP:conf/cvpr/GuoYYWL19} on Scene Flow datasets. As shown in Figure 4, light blue area takes up less area in the error map of our MFM method, which illustrates that our method has better prediction accuracy by decompose the matching task into multiple stage for high-resolution 3D cost volume directly.

Figure 2 and Figure 3 visualize the error maps of our method and other methods  on KITTI 2012 and KITTI 2015, respectively. The error area takes up less area in the error map of our MFM method, which also demonstrates the effectiveness of the proposed MFM mechanism.

The prediction of the peak position in the similarity distribution determines the accuracy of the disparity obtained through 3D cost volume refinement, which is the first and the most important phase of the disparity prediction task. We count the proportion of the points with peak position deviating from the ground truth by more than 1 pixel and 3 pixels, respectively. As shown in Figure 5, PSMNet, GANet, and GwcNet all have a larger number of error points than our MFM method, especially on the deviated more than 1 pixel figure. Well begun is half done. The remarkable advantage of our method shown in Figure 5 largely determines our final state-of-art performance.

\subsection{Detailed Analysis of Proposed Method}
\paragraph{Ablation Study.}
\begin{table}[t]
	\centering
	\caption{The ablative prediction results of different variants of MFM on Scene Flow datasets. B:Baseline, de:decouple, ms:\emph{Multistage Matching}, sma: \emph{Stages Mutual Aid}.}
	\begin{tabular}{ccccccl}
		\hline
		Model&EPE& \textgreater{}1px  & \textgreater{}3 px \\
		\hline
		Baseline &0.77&8.03\% &3.30\%\\
		\hline
		B+de&0.86&6.30\%&3.10\%\\
		\hline
		B+de+ms&0.74&5.18\%&2.66\%\\
		B+de+ms+sma&\textbf{0.66}&\textbf{4.95\%}& \textbf{2.50\%} \\
		\hline
	\end{tabular}
    \label{table3}
\end{table}
We conduct ablation studies to understand the influence of different designs in our proposed method. We design different runs on Scene Flow datasets and report the results in Table 4. First, GwcNet\cite{DBLP:conf/cvpr/GuoYYWL19} is adopted as our baseline. The "Baseline" refine the correlation features multiple times for a low-resolution 3D cost volume and the high-resolution 3D cost volume is obtained by linear interpolation. When we introduce the learning mechanism(Baseline+decouple) to learn all similarity scores from the last stage of the cost volume aggregation module, the $> 1px$ and $> 3px$ prediction error on Scene Flow datasets improve $1.73\%$ and $0.20\%$, respectively. However, the EPE drops slightly for the competition of simultaneously learning multiple similarity scores from one correlation feature, which may influence the prediction results on a large pixel error range. Then, we take account into the multistage decomposition of the learning task(Baseline+decouple +\emph{Multistage matching}), and achieve the state-of-the-art result on all accuracy metrics. This demonstrates that decoupling all similarity scores step by step from different 4D volume makes the task easier to learn. Finally, the \emph{Stages Mutual Aid} operation is added to ‘Baseline+decouple +\emph{Multistage matching}’. The performance is significantly improved, which demonstrates  the simplicity and effectiveness of the \emph{Stages Mutual Aid} module. Ablation experiments have verified that the proposed MFM scheme indeed learns the more accurate similarity scores by decomposing the task into multiple stages, thus effectively improves the prediction performance of the high-resolution disparity map.  

\section{Conclusion}
In this paper, we propose the Multistage Full Matching framework, which directly estimates high-resolution 3D cost volume from the low-resolution 4D volume by decomposing the matching task into multiple stages. First, a serial network is designed for sufficient cost aggregation and multistage high-resolution 3D cost volume estimation. Then, the  \emph{Stages Mutual Aid} is proposed to solve the unbalanced prediction of the multiple stages resulted by serial network. The last but the most important, our MFM scheme achieves state-of-the-art on three popular datasets.



\bibliographystyle{aaai21}
\bibliography{references}
\end{document}